\pdfoutput=1
\documentclass{article}

\usepackage[preprint,nonatbib]{neurips_2022}

\usepackage[numbers]{natbib}

\usepackage[utf8]{inputenc} %
\usepackage[T1]{fontenc}    %
\usepackage{textcomp}
\usepackage{hyperref}       %
\usepackage{url}            %
\usepackage{booktabs}       %
\usepackage{amsfonts}       %
\usepackage{nicefrac}       %
\usepackage{microtype}      %
\usepackage{xcolor}         %
\usepackage{todonotes}

\usepackage{lipsum}
\usepackage{caption}%
\usepackage[ruled,vlined,linesnumbered]{algorithm2e}  
\usepackage{algorithmic}

\usepackage[english]{babel}
\usepackage{xcolor}
\usepackage{xspace}
\usepackage{colortbl}
\usepackage{pifont}
\usepackage{subcaption}
\usepackage{wrapfig}
\usepackage{graphicx}
\usepackage[inline]{enumitem}
\usepackage{booktabs}
\usepackage{threeparttable}
\usepackage{times}
\usepackage[small, compact]{titlesec}
\usepackage[textfont={it}, font={small,bf}, skip=1pt]{caption}

\usepackage{multirow}
\usepackage{fancybox} %
\usepackage{amsmath}
\usepackage{amsthm}
\usepackage{comment}
\usepackage{array}
\usepackage{tabularx}
\usepackage{svg}

\usepackage{hyperref}
\hypersetup{
  colorlinks=true,      %
  linkcolor=blue,       %
  citecolor=magenta,    %
  filecolor=cyan,       %
  urlcolor=red          %
}
  
\usepackage{xspace}

\usepackage{tikz}
\usepackage{amsmath}
\usepackage{pgfplots}
\usepackage{hhline}
\usepackage{listings}

\definecolor{codegreen}{rgb}{0,0.6,0}
\definecolor{codegray}{rgb}{0.5,0.5,0.5}
\definecolor{codepurple}{rgb}{0.58,0,0.82}
\definecolor{backcolour}{rgb}{0.95,0.95,0.92}

\def\ie{{i.e.}}
\def\eg{{e.g.}}

\definecolor{codegreen}{rgb}{0,0.6,0}
\definecolor{codegray}{rgb}{0.5,0.5,0.5}
\definecolor{codepurple}{rgb}{0.58,0,0.82}
\definecolor{backcolour}{rgb}{0.95,0.95,0.92}

\def\name{Swan\xspace}

\newenvironment{denseitemize}{
	\begin{itemize}[topsep=2pt, partopsep=0pt, leftmargin=1.5em]
		\setlength{\itemsep}{2pt}
		\setlength{\parskip}{0pt}
		\setlength{\parsep}{0pt}
	}{\end{itemize}}

\newenvironment{denseenum}{
  \begin{enumerate}[topsep=2pt, partopsep=0pt, leftmargin=1.5em]
    \setlength{\itemsep}{2pt}
    \setlength{\parskip}{0pt}
    \setlength{\parsep}{0pt}
  }{\end{enumerate}}
    
\title{\name: A Neural Engine for Efficient DNN Training on Smartphone SoCs}

\author{%
  Sanjay Sri Vallabh Singapuram \\
  Department of Computer Science \\
  University of Michigan - Ann Arbor \\
  \texttt{singam@umich.edu} \\
  \And
  Fan Lai \\
  Department of Computer Science \\
  University of Michigan - Ann Arbor \\
  \texttt{fanlai@umich.edu} \\
  \And
  Chuheng Hu \\
  Department of Computer Science \\
  John Hopkins University \\
  \texttt{chu29@jhu.edu} \\
  \And
  Mosharaf Chowdhury \\
  Department of Computer Science \\
  University of Michigan - Ann Arbor \\
  \texttt{mosharaf@umich.edu} \\
}

\begin{document}

\maketitle

\begin{abstract}
The need to train DNN models on end-user devices (e.g., smartphones) is increasing with the need to improve data privacy and reduce communication overheads.
Unlike datacenter servers with powerful CPUs and GPUs, modern smartphones consist of a diverse collection of specialized cores following a system-on-a-chip (SoC) architecture that together perform a variety of tasks. 
We observe that training DNNs on a smartphone SoC without carefully considering its resource constraints can not only lead to suboptimal training performance but significantly affect user experience as well.
In this paper, we present \name, a neural engine to optimize DNN training on smartphone SoCs without hurting user experience.
Extensive large-scale evaluations show that \name can improve performance by $1.2-23.3\times$ over the state-of-the-art.
\end{abstract}

\section{Introduction}

Model training and inference at the edge are becoming ubiquitous for better privacy~\cite{fl-survey}, localized customization~\cite{mistify}, low-latency prediction~\cite{mistify} and etc. 
For example, Google~\cite{fl-google} and Meta~\cite{fl-meta} are running federated learning (FL) across potentially millions of end-user devices to train the model at the data source to mitigate privacy concerns in data migration; Apple performs federated evaluation and tuning of automatic speech recognition models on mobile devices \cite{fl-apple}; communication constraints like intermittent network connectivity and bandwidth limitations (\eg, car driving data) also necessitate the capability to run models closer to the user~\cite{focus}. 

Naturally, many recent advances for on-device model execution are focusing on optimizing ML models (\eg, model compression~\cite{deep-compression} or searching lightweight models~\cite{mobilenet}) or algorithm designs (\eg, data heterogeneity-aware local SGD in federated learning~\cite{fed-yogi}).
However, the execution engines that they are relying on and/or experimenting with are ill-suited for resource-constrained end devices like smartphones. 
Due to the lack of easily-extensible mobile backends, today's on-device efforts often resort to either traditional in-cluster ML frameworks (\eg, PyTorch~\cite{pytorch} or TensorFlow~\cite{tensorflow}) or operation-limited mobile engines (\eg, DL4J~\cite{dl4j}). 
The former does a poor job in utilizing available resources, while the latter limits which models to run.
Overall, existing on-device DNN training solutions are suboptimal in performance, harmful to user experience, and limited in capability.

Unlike cloud or datacenter training devices (i.e., GPUs), smartphones are constrained in terms of the maximum electrical power draw and total energy consumption; they cannot sustain peak performance for long. 
Modern smartphones use a system-on-a-chip (SoC) architecture with heterogeneous cores, each with different strengths and weaknesses.
When to use which core(s) to perform DNN training requires careful consideration of multiple constraints.
For example, one may want to use the low-performance, low-power core(s) for training to meet energy and power constraints.
However, this comes at the cost of longer training duration; in some cases, it may even be energy-inefficient due to longer duration outweighing the benefit of low-power execution.
It is, therefore, necessary to find a balance between low-latency and high-efficiency execution plans.
In short, we need a bespoke \emph{neural engine} for on-device DNN training on smartphone SoCs.

The challenges only increase as we start considering the dynamic constraints smartphones face that are typically not observed in datacenters. 
Smartphones today keep running a host of services, while prioritizing quick responses to user-facing applications. 
Because end users may actively be using a smartphone, the impact on these foreground applications must be minimal.
Running a computationally-intensive workload like DNN training can significantly degrade user experience due to resource contention. 
Existing proposals to offload training to unused cores cite{} are still just proposals without any available implementation.
At the same time, statically allocating cores to applications leads to resource underutilization.

In this paper, we propose \name, a neural engine to train DNN models on smartphones in real-world settings, while considering constraints such as resource, energy, and temperature limits without hurting user experience. 
Our key contributions are as follows: 
\begin{denseitemize}
\item \name is built within Termux, a Linux Terminal emulator for Android, and can efficiently train unmodified PyTorch models. 

\item We present and implement a resource assignment algorithm in \name to exploit the architectural heterogeneity of smartphone SoCs by dynamically changing the set of core(s) it uses to match on-device resource availability. 

\item We evaluate \name using micro- and macro-scale experiments. 
  Using the former, we show that \name reduces interference to foreground applications while improving local training performance. 
  Using the latter, we show that \name applied to smartphones participating in a federated learning setting can lead to global performance as well.

\end{denseitemize}

\section{Related Work}
\label{gen_inst}

\paragraph{On-Device Execution} 
Existing ML algorithms have made considerable progress to train the model on the edge. For example, in federated learning, FedProx~\cite{fed-prox}, FedYoGi~\cite{fed-yogi} and Fed-ensemble~\cite{fed-ensemble} reinvent the vanilla model aggregation algorithm, FedAvg~\cite{fed-avg}, to mitigate the data heterogeneity.
Oort~\cite{oort-osdi} orchestrates the global-scale FL clients, 
and cherry-picks participants to improve the time-to-accuracy training performance, while other advances are reducing network traffics~\cite{fetch-sgd}, enhancing client privacy via differential 
privacy~\cite{apple-dp, fl-hitter-dp}, personalizing models for different clients~\cite{mistify, fl-clustering}, and 
benchmarking FL runtime using realistic FL workloads (e.g., FedScale~\cite{fedscale} and Flower~\cite{flower}). 

On the other hand, recent execution frameworks, like Apple's CoreML \cite{coreml} for its mobile devices and Android's NNAPI \cite{nnapi}, offloads the inference to the mobile GPU or a Neural Processing Unit (NPU) to accelerate model inference. 
Deeplearning4J~\cite{dl4j} and PyTorch offer Java binaries to include with Android applications for on-device training, but they are not space-optimized (can be up to 400 MB), and they lack the capability to offload training to GPUs as well. 
Worse, these existing mobile engines require lots of engineering efforts to implement and experiment with new designs. 

\paragraph{Heterogeneity of Smartphone SoCs} 
A smartphone's application processor (AP) uses the same compute elements like a desktop computer, but draws lower power and has a smaller footprint. 
The CPU, GPU, memory, and other heterogeneous elements are packed into a single die known as a system-on-a-chip (SoC). 
ARM-based smartphones SoCs overwhelmingly dominate the smartphone market because of their energy efficiency as well as ease of licensing \cite{46755}.

\begin{figure}
	\centering
	\begin{subfigure}[b]{0.45\textwidth}
		\centering
		\includegraphics[width=\textwidth]{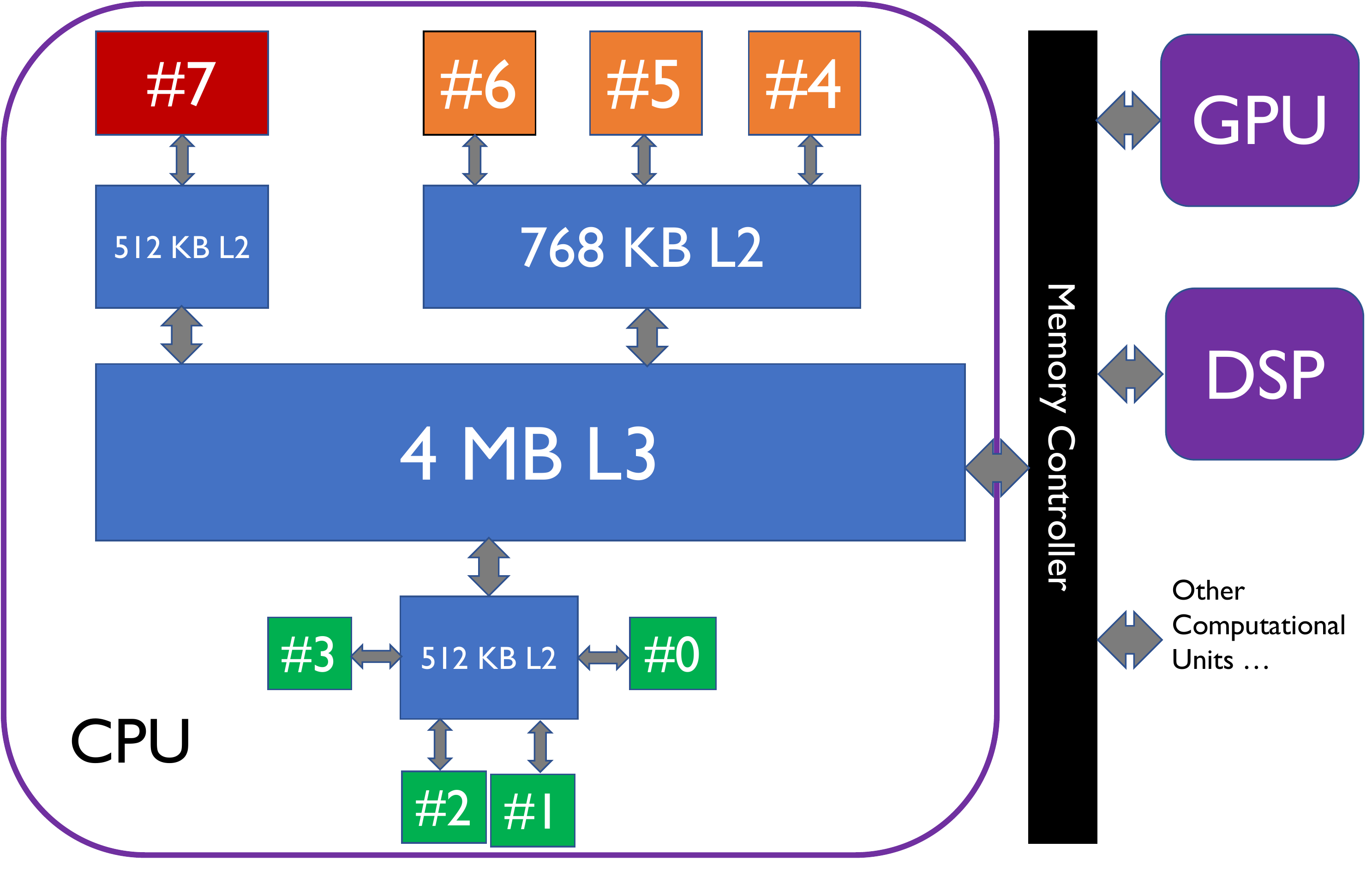}
		\caption{SD865's Heterogenous SoC Architecture}
		\label{fig:sg865-arch}
	\end{subfigure}
	\hfill
	\begin{subfigure}[b]{0.53\textwidth}
		\centering
		\includegraphics[width=\textwidth]{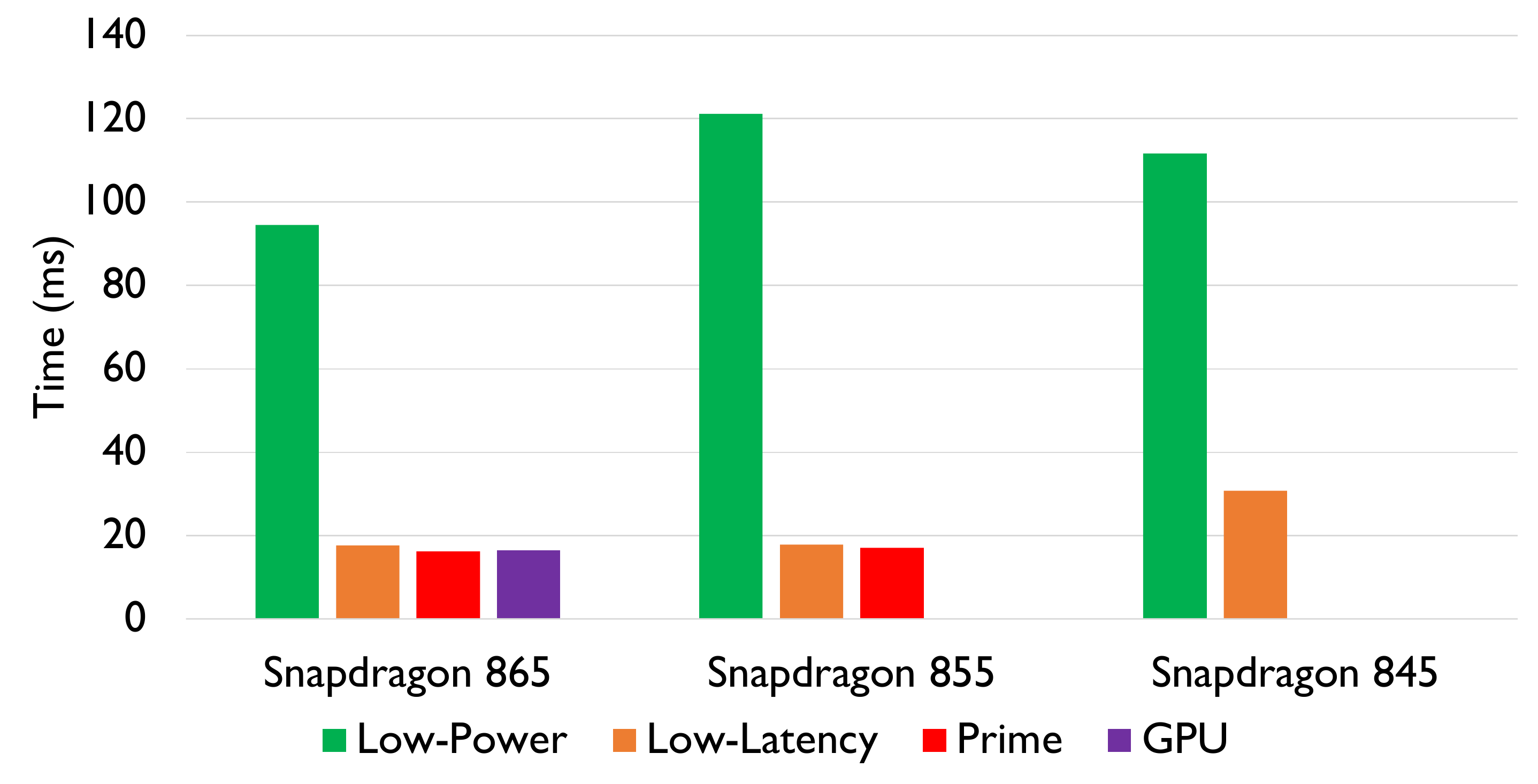}
		\caption{Per-Core 512x512 Matmul Performance across SoCs}
		\label{fig:matmul_core}
	\end{subfigure}
	   \caption{Heterogeneity in Smartphone SoCs}
	   \label{fig:hetero-soc}
\end{figure}

Available compute cores in smartphone SoCs can vary widely. 
For example, the Snapdragon SD865 SoC shown in Figure~\ref{fig:sg865-arch} has four low-powered cores (\#0-\#3) and four cores optimized for low latency (\#4-\#7) in addition to GPU and DSP.
One of its low-latency cores ("Prime" core \#7) is overclocked for even higher performance. 
Typically, all low-latency cores are turned off when the phone is idle to save battery.
The differences in performance characteristics of these cores are demonstrated by their time taken to multiply two 512 X 512 matrices in Figure~\ref{fig:matmul_core}, and also compares it to the performance of the entire GPU in Snapdragon 865.

\paragraph{Android vs. Linux Distros} 
The Android operating system is based on the Linux kernel, and therefore implements many of the Linux system calls and the directory structure. 
Unlike many Linux distros (e.g., Ubuntu), user-space applications are sand-boxed for security reasons \cite{sandbox}, \ie they cannot access files related to system-level information (e.g. /proc) related to other processes or overall information like CPU load. 
This makes it impossible to accurately gauge system-level information that could be used toward intelligent scheduling. 
One way to get around the sand-boxing is to ``root'' the Android device for user-space applications to gain access to system information, but that comes with the cost of possibly making the device unusable (bricking) and/or making the user data vulnerable to malicious applications \cite{casati2018dangers}.

\section{Motivation}
\label{motivation}

\begin{figure}[!htbp]
	\centering
	\begin{subfigure}{0.45\textwidth}
		\includegraphics[width=\linewidth]{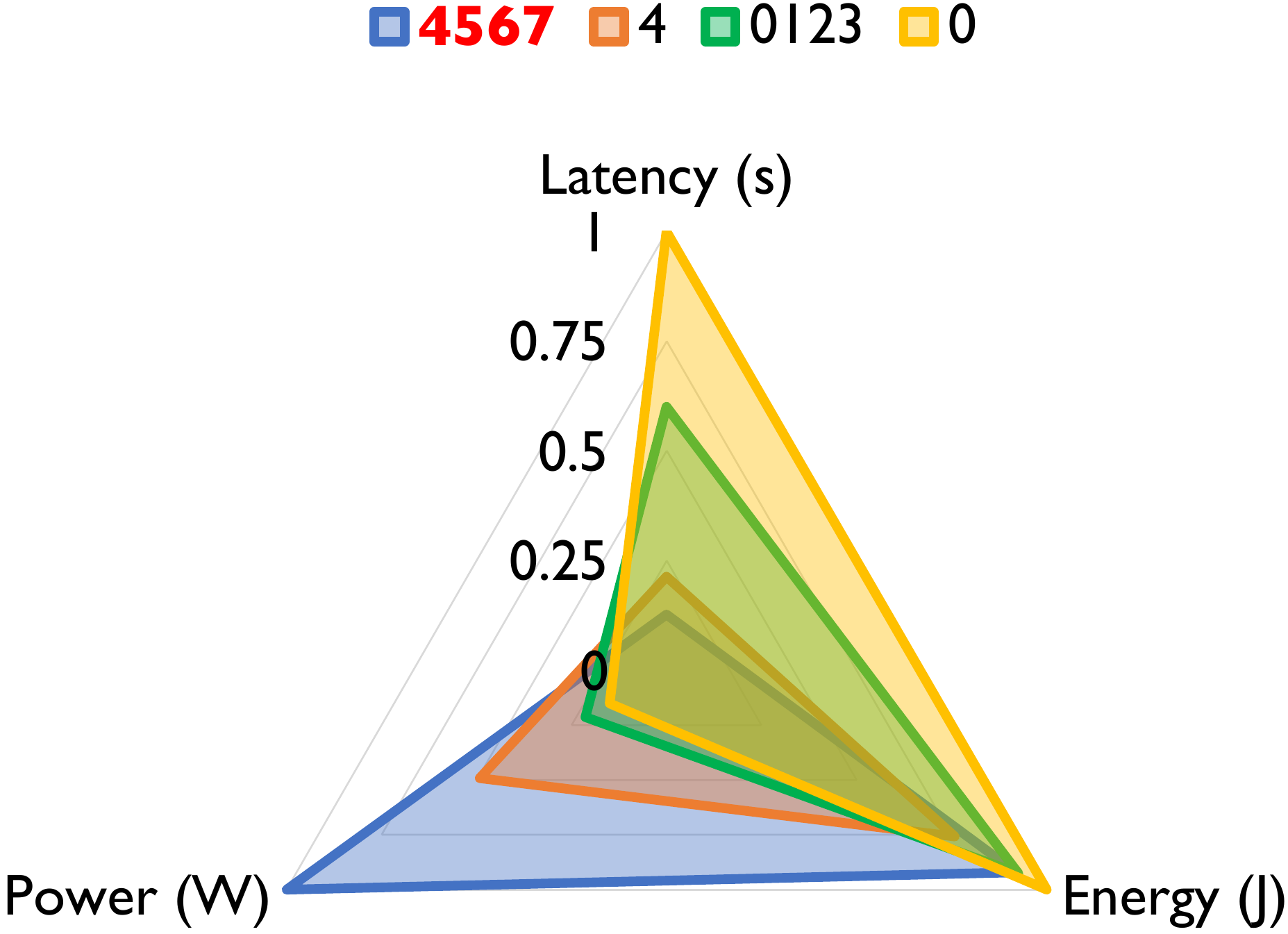}
		\caption{Resnet34 on Pixel 3}
		\label{fig:pxl3-resnet-all}
	\end{subfigure}
	\begin{subfigure}{0.45\textwidth}
		\includegraphics[width=\linewidth]{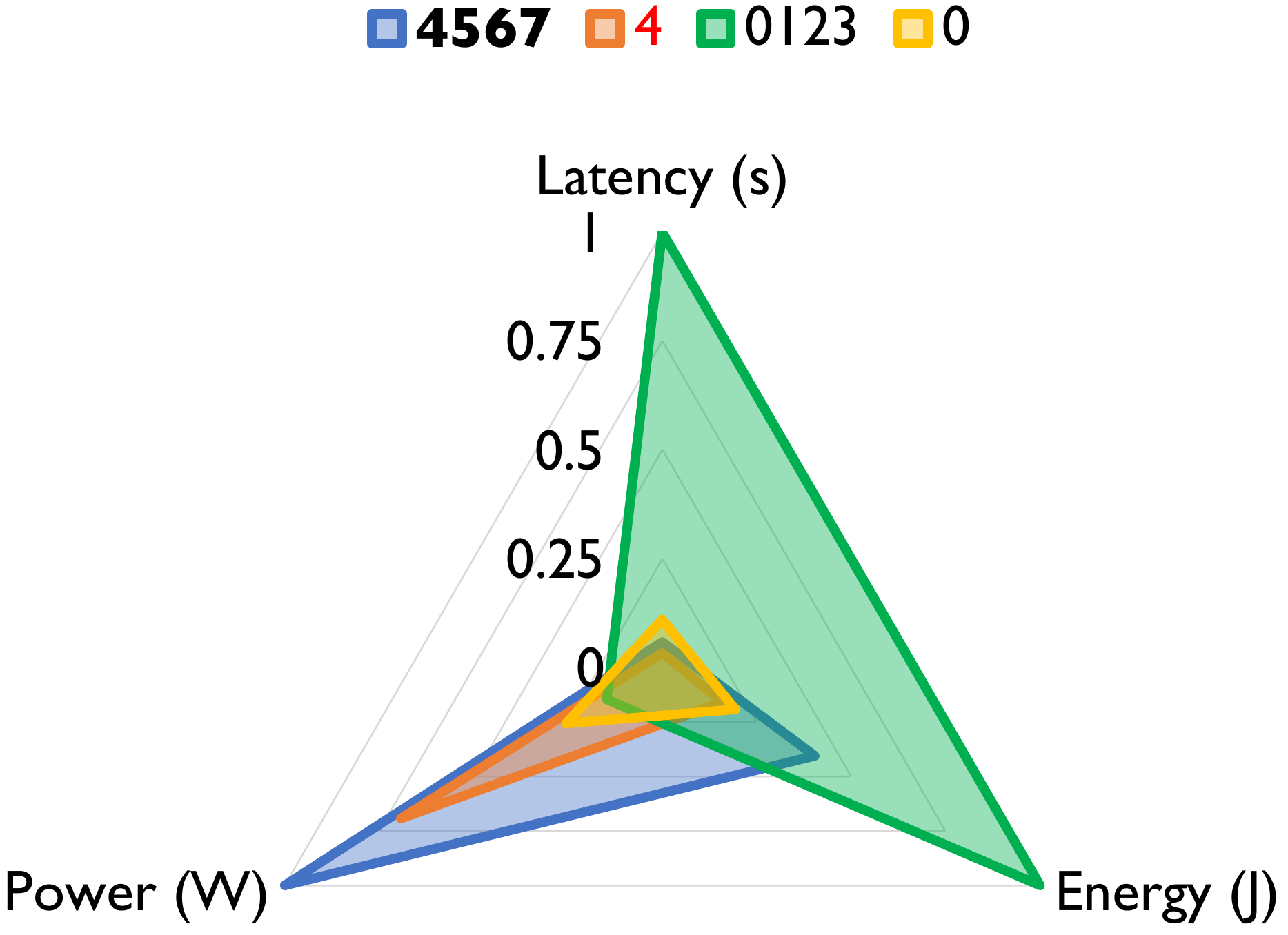}
		\caption{ShuffleNet on Google Pixel 3}
		\label{fig:pxl3-shufflenet-all}
	\end{subfigure}
	\caption{Relative comparison of avg. latency, energy and power usage per core-combination}
	\label{fig:resusage-spider-chart}
\end{figure}

\subsection{The Impact of SoC and Model Architectures on On-Device Training}
The choice of the cores on an SoC when training a DNN model, as well as the model architecture itself significantly, affects performance across multiple dimensions such as time taken (latency), peak power drawn, and total energy consumed.

Figure~\ref{fig:pxl3-resnet-all} details the resource usage to train Resnet34 on Pixel 3 using PyTorch, on a variety of CPU core combinations. 
PyTorch uses a \emph{greedy} strategy to pick as many threads as there are low-latency cores. 

The fastest choice to train the network is to use all the low-latency cores (i.e., \texttt{4567}), with the speed reducing as the number of cores reduce and/or less powerful cores are used. 
This suggests that the workload benefits from scaling. 

In contrast, the most energy-efficient choice is to involve any one of the low-latency cores (i.e., 4, 5, 6, or 7). 
This brings up an interesting observation that \emph{low power usage does not translate to low energy usage}: while combinations involving the low-power cores (i.e., 0--3) are always more power-efficient, they do not tend to be more energy-efficient. 
Lower power leads to slower execution, thus increasing the total energy spent over time.

These observations are not universal, however.
For example, training ShuffleNet on the Pixel 3 (Figure~\ref{fig:pxl3-shufflenet-all}) results in using one of the low-latency cores as both the fastest and most energy-efficient choice. 
The reason for this apparent drawback of scaling is due to the presence of depth-wise convolution operations, which are more memory-intensive than standard convolution operations \cite{qin2018diagonalwise, zhang2018shufflenet}. 
Multiple threads running memory-intensive operations make them compete for the cache, leading to cache-thrashing and reducing overall performance. 
Using just one thread allows the cache to be used in an exclusive manner. 
This is a known issue that has been addressed in GPUs \cite{qin2018diagonalwise} and Intel CPUs \cite{mkldnn-dwconv}, but is yet to be addressed for ARM CPUs. 
In the absence of a pre-optimized training backend, it is necessary to customize the execution for every DL model and every smartphone at hand.

\subsection{The Impact of On-Device Training on User Experience}

\begin{wrapfigure}{r}{0.5\textwidth}
	\begin{center}
		\includegraphics[width=0.48\textwidth]{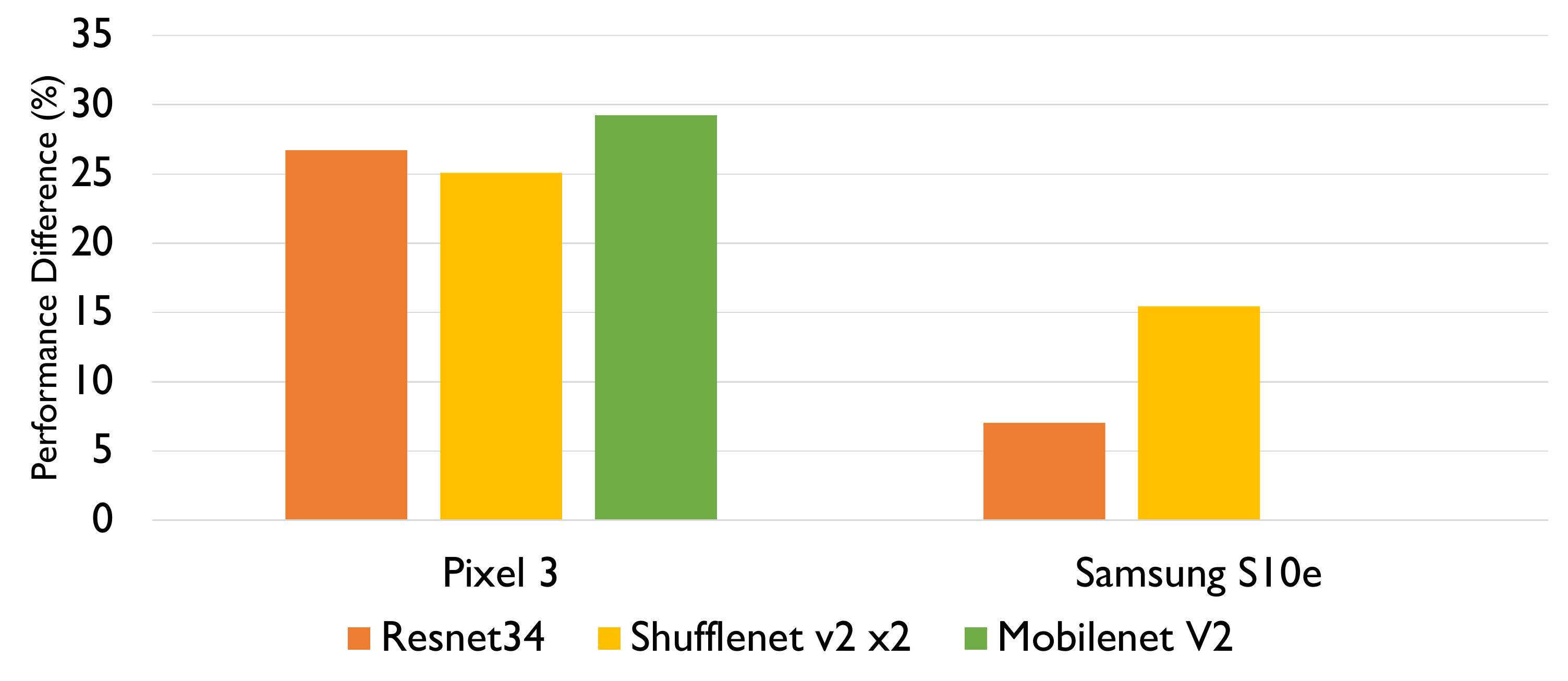}
	\end{center}
	\caption{Background training significantly reduces foreground PCMark benchmarking score.}
	\label{fig:pcmark-interference-bground}
\end{wrapfigure}

Scheduling on-device training only when the device is detected to be idle has the benefit of reducing adverse impact on user experience since training is resource-intensive \cite{fl_google, ryffel2018generic}. 
The adverse user experience can manifest as slower responses to user interactions or delayed video playback. 
We can measure this impact by running a benchmark that is representative of real-world usage, like PCMark Work 3.0 \cite{pcmark-android}, with and without the training processes running in the background. 
As shown in Figure~\ref{fig:pcmark-interference-bground}, the training process does have an impact on the benchmark score, with the less performant Pixel 3 being impacted more than the Samsung S10e. 
With a majority of Android applications only using 1--2 threads \cite{7095808}, this presents an opportunity to exploit other CPU cores that are either under lower load or are being used by low-priority background services, enabling the training to run even the phone is being used.

\section{\name}
\label{sec:design}

\name is a neural engine for on-device DNN training on smartphone SoCs that improves the performance and energy efficiency of training while minimizing the impact on user experience. 
This leads to improved performance for mobile applications locally as well as quicker model convergence for distributed applications such as federated learning. 
Figure~\ref{fig:swan-design} outlines the overall architecture of \name.

\subsection{Design Overview}

At its core, \name explores combinations of CPU cores as execution choices to optimize for training time and also as alternative choices for the execution to migrate to when training interferes with foreground applications. 
Migrating training to fewer cores or to cores that are used by background processes relinquishes compute resources for the user-facing applications. 
To this end, we utilize the heterogeneity in smartphone SoCs to provide many execution choices to ensure that the device can continue training under a wider range of resource constraints. 
\name infers interference to/from other applications without rooting the device and does not need invasive power monitors to measure the energy expenditure of on-device training, thus enabling large-scale deployment on Android devices.

\paragraph{Standardized Interface} 
We intend the communication interface of \name at the client to follow the existing standard (i.e., client implements \texttt{isActive} and \texttt{run\_local\_step}) in order to work seamlessly with existing distributed solutions such as federated learning server-client frameworks (e.g., PySyft \cite{pysyft}). 
This is also to reduce the possibility of introducing unintentional privacy leaks by deviating from the standardized client-coordinator interfaces.

\begin{figure}
	\centering
	\begin{subfigure}[b]{0.54\textwidth}
		\centering
		\includegraphics[width=\linewidth]{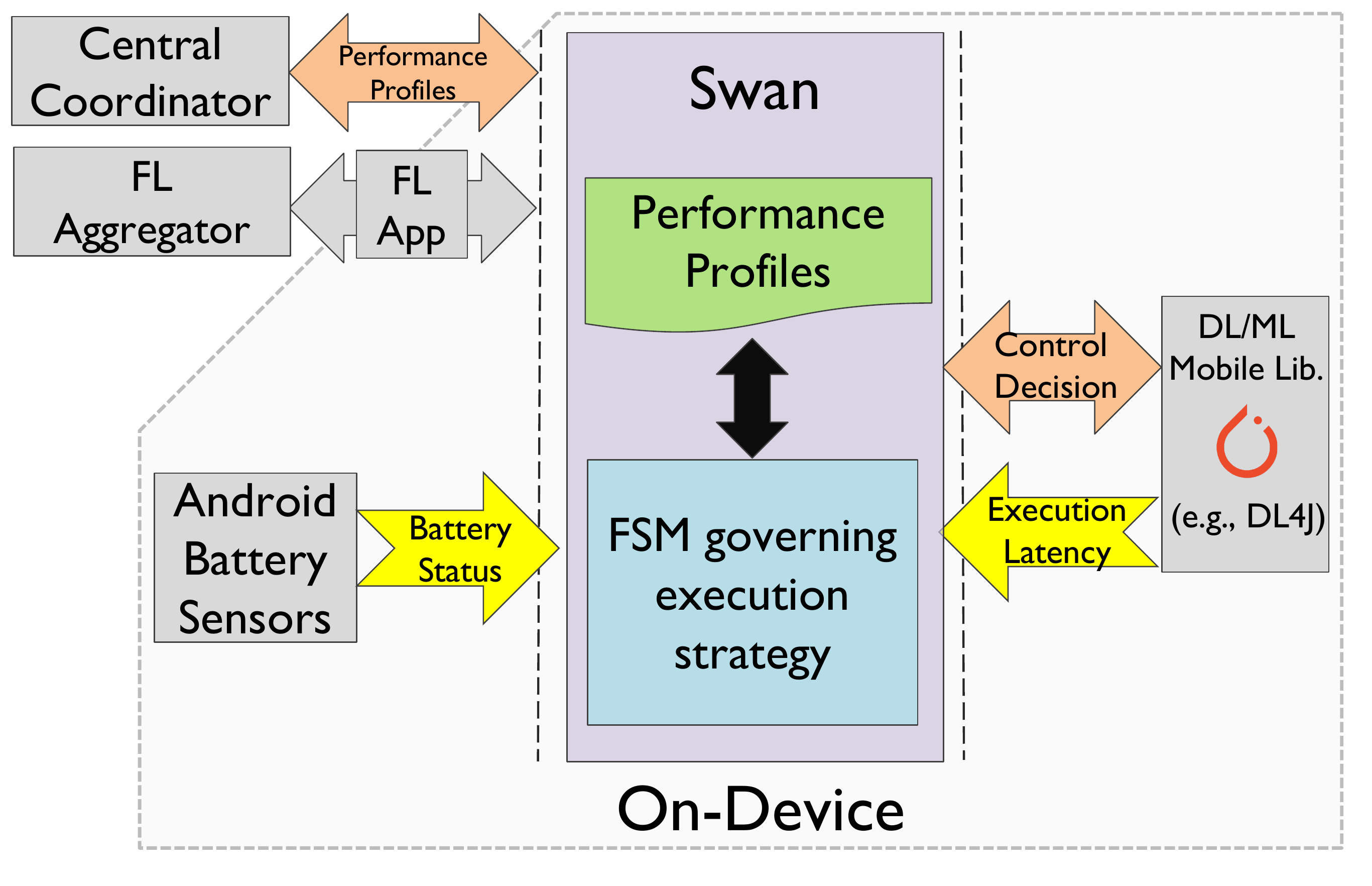}
		\caption{Architecture of \name}
		\label{fig:swan-arch}
	\end{subfigure}
	\hfill
	\begin{subfigure}[b]{0.41\textwidth}
		\centering
\begin{lstlisting}[language=Python]
for _ in range(num_batches):
    batter_status = swan.get_battery()
    if batter_status != "charging" or batter_status < MIN_BAT_THRESHOLD:
        break
    
    latency = train(execution_choice)
    # Detect interference
    if latency > latencies[execution_choice] * 1.1:
        # Downgrade Execution Choice
        execution_choice -= 1
        grade_time = time.now()
    # Explore to upgrade choice
    elif time.now() - grade_time > _2_minutes:
        execution_choice += 1
    
    swan.migrate(execution_choice)
\end{lstlisting}
		\caption{Control Loop}
        \label{lst:control-loop}
	\end{subfigure}
	\caption{Design of \name}
	\label{fig:swan-design}
\end{figure}

Here, we summarize the sequence of steps required to involve a smartphone under \name and go deeper into each step in the following subsections:

\begin{enumerate}
	\item \textbf{Monitoring}: 
	After installing \name on the device, \name monitors the battery state to decide on a training request based on whether it has completed its execution choice exploration and the device being idle. 
	\name also declines the request if the battery is above 35\textdegree C, which is required in real to prevent battery life reduction \cite{ma2018temperature, huawei-batt} and thermal pain\cite{egilmez2015user}. 
	When not servicing requests, \name monitors the rate of battery charge loss to determine the background services' power usage.
		
	\item \textbf{Exploring Execution Choices}: 
	Upon receiving a training request, \name picks one of the unexplored choices to determine its resource usage (\ie, energy, power) and performance. 
	\name explores only if the device is idle and the battery is discharging, since the amount of discharge is related to the energy used by the processor. 
	Ensuring that the phone was idle simplifies our energy measurement by attributing the energy usage only to the training and the background services.
	
	\item \textbf{On-Device Training}: 
	Similar to real FL deployments~\cite{fl-google}, \name accepts a training request if the battery is charging, 
	while it can admit the training too once the device battery is above a minimum level. This is because \name can adaptively activate the execution to prevent lower battery levels. 
	However, this could potentially be side-stepped by introducing users to provide their preference in future. 
	It then uses the performance profiles of execution choices to dynamically migrate the execution based on inferred interference (Figure~\ref{lst:control-loop}).
\end{enumerate}

While we have implemented \name as a userland service in the Android, we envision that it can be integrated to Android OS itself as a neural engine service that can be used by all applications running on the device. 

\subsection{Exploring Execution Choices}
\label{sec:exec-explore}

In order to accommodate varying compute resource availability during training, we explore running the training task on different combinations of compute units. 
These combinations can be a selection of CPU cores, or other execution units like the mobile-embedded GPU. 
Since the PyTorch execution backend we use is implemented only for CPUs, we limit the exploration to a combination of CPU cores, but design our system to be agnostic to the execution choice to be able to include other execution units in the future (Figure~\ref{fig:swan-arch}). 
Each combination is benchmarked by training on a small number of batches (with a minimum specified by the request, and the rest running on a copy of the model) and amortizing the resource usage for one local step. 
In Appendix~\ref{app:implementation}, we delve deeper into the state-space of execution choices we explore. 
The device can perform this exploration in a work-conserving manner, by participating in model training while benchmarking.

The exploration process can be further amortized by leveraging the central aggregator(s) present in distributed learning systems. 
For example, the central aggregator in federated learning can distribute the list of execution choices to explore amongst devices of the same model, thereby accelerating the exploration process and preventing each user to bear the brunt of exploring all the execution choices. 
Once all choices are explored and the performance profiles reported back to the coordinator, new devices with \name installed benefit from this available knowledge by skipping the exploration step altogether.

\subsection{Making the Execution Choice}

Once the profiling is done, we need to prioritize the profiles in such a way that the fastest profile is picked under no interference and ``downgrading'' to a profile during interference leads to relinquishing compute in favor of the interfering applications.

To this end, we first sort all profiles in the order of increasing the expected training time. 
In order to ensure that the ``downgrade'' choice is able to relinquish compute, we define a cost for each execution choice. 
The Android source code offers an insight into the scheduling of applications' threads onto processor cores based on their priority \cite{android-cpusets}. 
Our scheduling policy prefers to dedicate the faster cores to the application being currently used (i.e., the foreground application) and other foreground-related services. 
This provides us a way to identify a total order between execution choices. These rules include,

\begin{denseenum}
	\item Using more cores of the same type is costlier (e.g. cost['4567'] > cost['4']).
 
	\item Using any number of low-latency cores is costlier than using any number of low-power (high-latency) cores (\eg cost['4'] > cost['0123'])
	
	\item For devices with a Prime core, the Prime cores are considered costlier than low-latency cores (\eg cost['47'] > cost['45']) since relinquishing choice \texttt{47} for \texttt{45} allows other applications to use the Prime core.
\end{denseenum}

For example, following these guidelines, the cost order for Pixel 3 would be "4567" > "456" > "45" > "4" > "0123" > "012" > "01" > "0".

We can then prune choices that cost more than the choices that precede them, \ie, effectively removing choices that present no viable tradeoff. 
For example, while choosing \texttt{4-7} over \texttt{4} to train Resnet34 on Pixel3 presents a tradeoff between latency and energy efficiency since cost['4567'] > cost['4'] but '4567' is faster than '4' (Figure~\ref{fig:pxl3-resnet-all}). 
However, choosing \texttt{4-7} to train ShuffleNet worsens both latency and energy efficiency compared to \texttt{4}, which is also costlier than \texttt{4} since \texttt{4567} uses more cores (Figure~\ref{fig:pxl3-shufflenet-all}). Pruning thus reduces the chance of \name interfering with other applications.

The performance profile of every DL model-execution choice combination can in turn inform the design of the DL model based on whether it is able to maximize the utilization of all compute resources. 
For example, ShuffleNet needs to scale with more cores by addressing bottlenecks in the multi-core scenario.

\section{Evaluation}

We evaluate \name in real-world settings and in a simulated setting of federated learning to gauge its large-scale impact in distributed settings, by training three deep-learning models on CV and NLP datasets.

\subsection{Methodology}

\paragraph{Experimental Setup} We benchmark the energy usage and latency for each DL model on 5 mobile-devices: Galaxy Tab S6, OnePlus 8, Samsung S10e, Google Pixel 3 and Xiaomi Mi 10, to obtain their performance profiles. We detail our benchmarking methodology in Appendix \ref{app:perf-bench}.

In order to evaluate \name across a large scale of devices, we use an open-source FL benchmark, FedScale~\cite{fedscale, fedscale-repo}, to emulate federated model training using 20 A40 GPUs with 32 GB memory, wherein we replace FedScale system trace with our specific device trace. The trace is a pre-processed version of the GreenHub dataset \cite{green-hub} collected from 300k smartphones, and its pre-proccessing is detailed in Appendix \ref{app:preproc}.

To measure impact on user-experience, we compare the PCMark for Android \cite{pcmark-android} benchmark score obtained with and without running the local training on a real-device. We specifically chose PCMark due to its realistic tests (\eg, user web browsing) over many other benchmarks that simply stress test mobile componenets including the CPU \cite{passmark-android, geekbench-android}. The overall score is calculated from scores of individual tests that were impacted by local training. \name dynamically chooses execution choices to move away from cores under contention.

\paragraph{Datasets and Models} We run two categories of applications with real-world datasets of different scales, detailed in Table \ref{tab:data-stats}.
\begin{denseitemize}
	\item \emph{Speech Recognition:} The ResNet34 model is trained on the small-scale GoogleSpeech dataset to recognize a speech command belonging one of 35 categories.
	\item \emph{Image Classification:} The MobileNet and ShuffleNet models are trained on 1.5 million images mapped to 600 categories on the OpenImage dataset.
\end{denseitemize}

\begin{wrapfigure}{r}{0.5\textwidth}
	\centering
	\scalebox{0.9}{
	\begin{tabular}{ccc}
	\hline
	\textbf{Dataset} & \textbf{\# of Clients} & \textbf{\# of Samples} \\ [.8ex] \hline
	Google Speech\cite{google-speech} & 2,618 & 105,829  \\ [.8ex]
	OpenImage \cite{openimg} & 14,477 & 1,672,231 \\ [.8ex] \hline
	\end{tabular}
	}
	\captionof{table}{Statistics of the dataset in evaluations.}
	\label{tab:data-stats}
\end{wrapfigure}

\paragraph{Parameters} The minibatch size is set to 16 for all tasks. The training uses a learning rate of 0.05 combined with the SGD optimizer. The baseline uses the execution choice defined by PyTorch that greedily picks as many threads there are low-latency cores in the device's SoC configuration, while \name picks the fastest execution choice amongst all that were explored. We use the Fed-Avg \cite{fed-avg} averaging algorithm to combine model updates.

\paragraph{Real-world energy budget} 
Many of the works proposed in FL do not factor in device failures that can be caused due to the unavailability of energy, effectively assuming an infinite energy budget. \cite{cho2022flame} uses an extremely conservative and static energy budget, effectively assuming that a device will never re-charge and replenish the extra energy used by FL. Accurately estimating the amount of extra energy delivered by the charger when running FL is also difficult since charging speeds vary according to their power output and the charging speed throttling to reduce battery wear. To simplifly the energy modeling of the charger while accounting for its presence, we fix the amount of energy delivered by the charger and energy usage of device usage on a daily basis, which is unique each device, thereby not assuming an infinite energy budget nor a static budget. We keep track of the energy used by FL, called the energy loan. The device is considered to be unavailable if the device ends up reaching its critical battery level if the energy loan were to reflect on the battery level from the device trace.

\paragraph{Metrics} For the local evaluation, we want to reduce execution latency while increase energy efficiency. For the large-scale FL simulation, we want to reduce the training time to reach target accuracy, while reducing energy usage. We set the target accuracy to be the highest achievable accuracy by either the baseline or \name.

\subsection{Local Evaluation}

\begingroup
\begin{table}[]
\centering
	\setlength{\tabcolsep}{6pt} %
    \renewcommand{\arraystretch}{1.5} %
	\small 
	\begin{tabular}{ccccccc}
	\hline
	\multirow{2}{*}{Device} & \multicolumn{3}{c}{Speedup}       & \multicolumn{3}{c}{Energy Efficiency} \\ \cline{2-7} 
						   & Resnet34 & ShuffleNet & MobileNet & Resnet34   & ShuffleNet  & MobileNet  \\ \hline
	Galaxy Tab S6          & 1.9$\times$     & 21$\times$        & 14.5$\times$     & 1.9$\times$       & 12.2$\times$       & 9.4$\times$       \\ \hline
	OnePlus 8              & 2.1$\times$     & 17$\times$        & 13.9$\times$     & 2.4$\times$    & 8.5$\times$     & 7.5$\times$    \\ \hline
	Google Pixel 3         & 1$\times$        & 1.8$\times$    & 1.6$\times$   & 1$\times$          & 1.8$\times$     & 2.3$\times$    \\ \hline
	Samsung S10e           & 1.9$\times$  & 39$\times$    & 31.8$\times$    & 2.1$\times$    & 39$\times$     & 17.4$\times$    \\ \hline
	Xiaomi Mi 10           & 2.1$\times$  & 17.2$\times$    & 14$\times$   & 2.2$\times$    & 7.8$\times$     & 5.8$\times$     \\ \hline
	\end{tabular}
	\vspace{.2cm}
	\caption{Local Speedups and Energy Efficiency Improvements over baseline.}
	\label{tab:local-improvements}
\end{table}
\endgroup

\paragraph{\name improves on-device execution latency and energy cost.} 
As shown in Table~\ref{tab:local-improvements}, while the baseline's heuristic is tied with \name only for the case of Resnet34 running on the Pixel 3 device, \name reduces training latency and energy expenditure by 1.8 - 39$\times$ for all other model - device combinations. The Shufflenet and Mobilenet models experience the most improvements due to higher latencies experienced due to cache-thrashing (Section~\ref{motivation}). \name finds faster execution choices that also reduce energy since the device operates at higher power states for a shorter period, leading to lower energy usage.

\begin{table}
	\centering
	\begingroup
		\setlength{\tabcolsep}{6pt} %
		\renewcommand{\arraystretch}{1.5} %
		\small
		\begin{tabular}{|l|l|l|l|}
			\hline
			Device         & Baseline   & \name      \\ \hline
			Galaxy Tab S6  & -10.2 \%   & -5.8 \%    \\ \hline
			OnePlus 8      & -12.5 \%   & 0 \%       \\ \hline
			Google Pixel 3 & -27 \%     & -3.1 \%    \\ \hline
			Samsung S10e   & -11.2 \%   & 0 \%       \\ \hline
			\end{tabular}
			\vspace{.2cm}
		\caption{Effect of interefence on PCMark Score while executing training in the background.}
		\label{tab:pcmark-eval}
\endgroup
\end{table}

\paragraph{Improvements in user-experience} From Table~\ref{tab:pcmark-eval}, we see that \name improves the user-experience according to the PCMark benchmark. The improvement is particularly stark for the Google Pixel 3 device since it is the lowest-end device amongst all, which is particularly important to enable on-device training since low-end devices make up a the majority of smartphones~\cite{46755}. The results also present implications for future mobile process scheduling algorithms to be more tighlty coupled to consider altering the number of threads and affinity for better load balancing. The Xiaomi Mi 10 being a high-end device, did not experience any impact of the background training.

\subsection{Large Scale evaluation}

\begin{figure}[h]
	\begin{subfigure}{0.49\textwidth}
		\includegraphics[width=\textwidth]{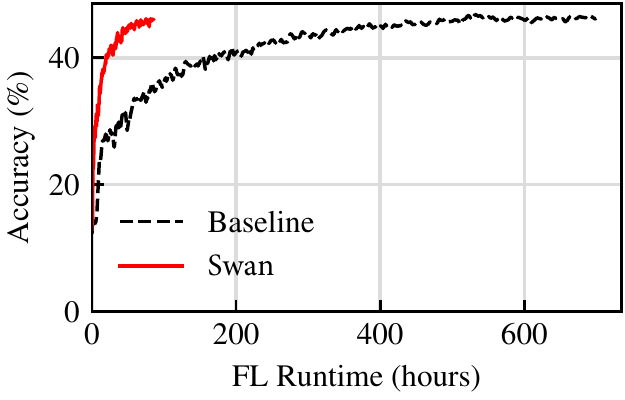}
		\caption{Time to accuracy performance.}
		\label{fig:shufflenet_tta}
	\end{subfigure}
	\hfill
	\begin{subfigure}{0.49\textwidth}
		\includegraphics[width=\textwidth]{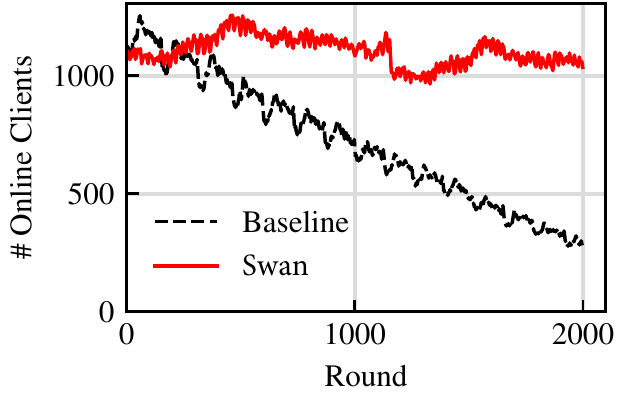}
		\caption{Number of clients online per round.}
		\label{fig:shufflenet_clients}
	\end{subfigure}
	\caption{Federated Training of ShuffleNet-V2.}
\end{figure}

\begin{figure}[h]
	\begin{subfigure}{0.49\textwidth}
		\includegraphics[width=\textwidth]{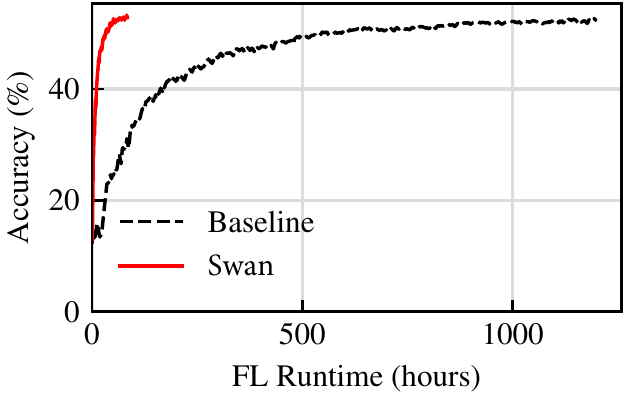}
		\caption{Time to accuracy performance.}
		\label{fig:mobilenet_tta}
	\end{subfigure}
	\hfill
	\begin{subfigure}{0.49\textwidth}
		\includegraphics[width=\textwidth]{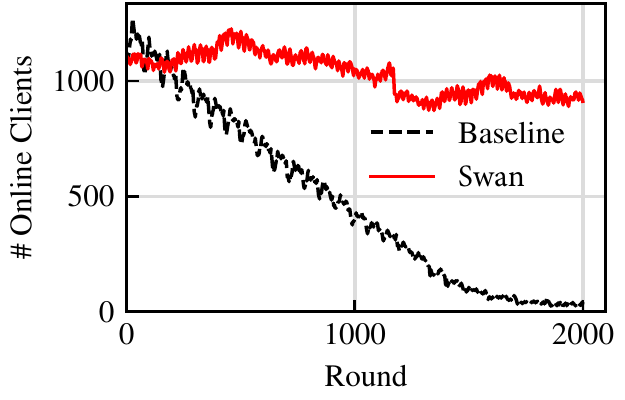}
		\caption{Number of clients online per round.}
		\label{fig:mobilenet_clients}
	\end{subfigure}
	\caption{Federated Training of MobileNet-V2.}

	\label{fig:acc-vs-time}
\end{figure}

\begin{figure}[h]
	\centering
	\begin{subfigure}{0.49\textwidth}
		\includegraphics[width=\textwidth]{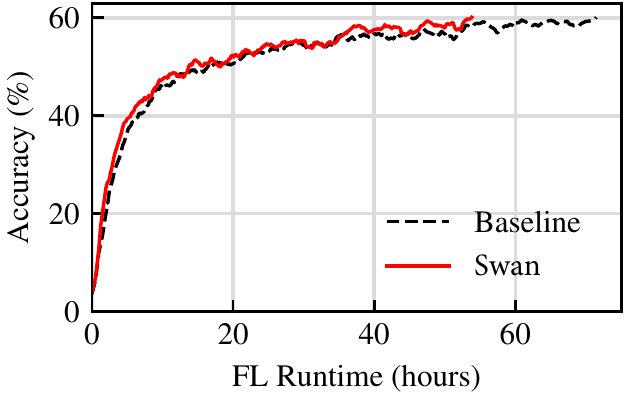}
		\caption{Time to accuracy performance.}
	\end{subfigure}
	\hfill
	\begin{subfigure}{0.49\textwidth}
		\includegraphics[width=\textwidth]{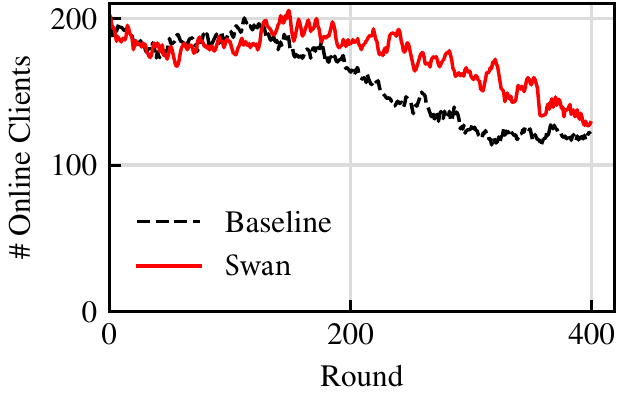}
		\caption{Number of clients online per round.}
	\end{subfigure}
	\caption{Federated Training of Resnet34.}
    \label{fig:resnet-perf}
\end{figure}

\paragraph{\name improves time to accuracy and energy efficiency for federated learning.}

\begin{table}[t]
	\setlength{\tabcolsep}{6pt} %
    \renewcommand{\arraystretch}{1.5} %
	\small
	\begin{tabular}{cclcccl}
		\hline
		\multirow{2}{*}{Task} &
		  \multirow{2}{*}{Dataset} &
		  \multirow{2}{*}{Model} &
		  \multirow{2}{*}{Target Acc.} &
		  \multirow{2}{*}{Speedup} &
		  \multirow{2}{*}{Energy Eff.} \\
										&                                                &                              &                &        &      &                 \\ \hline
		\multirow{2}{*}{Classification} & \multirow{2}{*}{OpenImage \cite{openimg}}      & MobileNet \cite{mobilenet}   & 52.8\%  & 23.3$\times$ & 7.0$\times$ \\ \cline{3-7} 
										&                                                & ShuffleNet \cite{zhang2018shufflenet} & 46.3\% & 6.5$\times$   & 5.8$\times$ \\ \hline
		Speech Recognition              & Google Speech \cite{google-speech}             & ResNet-34 \cite{resnet}      & 60.8\% & 1.2$\times$   & 1.6$\times$ \\ \hline
	\end{tabular}
	\vspace{.2cm}
	\caption{Summary of improvements on time to accuracy and energy usage in large-scale evaluation. }
	\label{table:e2e-perf}
\end{table}

We next dive into \name's benefit in large-scale FL training setting, where we train the ShuffleNet~\cite{zhang2018shufflenet}, MobileNet~\cite{mobilenet}, and ResNet-34~\cite{resnet} model in the image classification task and speech recognition task. 
Table~\ref{table:e2e-perf} summarizes the overall time-to-accuracy improvement, and Figure~\ref{fig:shufflenet_tta}-\ref{fig:mobilenet_tta} report the corresponding training performance over time. 
We notice that \name's local improvements with lower latency lead to faster convergence rates (1.2-23 $\times$) when applied to large-scale federated learning. \name improves energy efficiency by (1.6-7 $\times$) which directly translates to a higher number of devices that are available to perform training, unlike the baseline which steadily loses devices with every passing round due to exhausting the energy budget (Figure~\ref{fig:shufflenet_clients} and Figure~\ref{fig:mobilenet_clients}). Having more online devices for longer helps adapt the model to newer training data. Figure~\ref{fig:resnet-perf} reports the ResNet-34 performance on the speech recognition task. As expected, due to the small number of clients that can be used in this dataset, \name and baseline achieve comparable time-to-accuracy performance. However, \name achieves better energy efficiency by adaptively picking the right execution choice.

\section{Conclusion}
\label{sec:conclustion}

The need to train DNN models on end-user devices such as smartphones is only going to increase with privacy challenges coming to the forefront (e.g., recent privacy restrictions in Apple iOS and upcoming changes to Google Android OS).
Unfortunately, there is no existing solution that takes into account the challenges in DNN training computation on smartphone SoCs.
In this paper, we propose \name, to the best of our knowledge, the first neural engine specialized to efficiently train DNNs on Android smartphones.
By scavenging available resources, \name ensure faster training times and lower energy usage across a variety of training tasks, which leads to improvement for distributed training scenarios such as federated learning.
We believe that while this is only a first step, \name will enable further research in this domain and enable future researchers and practitioners to build on top its toolchain. 

\paragraph{Societal Impacts and Limitations}
We expect \name to be a standardized mobile execution engine for ML model deployment, which can facilitate today's ML research and industry. However, the potential negative impact is that \name might narrow down the scope of future papers to the PyTorch code that have been included so far. In order to mitigate such a negative impact and limitation, 
we are making \name open-source, and will regularly update it to accommodate to diverse backends based on the input from the community.

{
\bibliographystyle{plain}
\bibliography{swan}
}

\pagebreak 
\appendix

\section{Data Pre-Processing}

\subsection{Monitoring and modeling resource usage}

Due to the inherent scalability and logistical issues associated with collecting the resource usage data of smartphones, we emulate the background resource usage logging mentioned in Section~\ref{sec:design} by using a system trace dataset provided by GreenHub \cite{green-hub}. This data was collected by a background app logging the usage of various resources, resulting in collecting 50 million samples from 0.3 million Android devices that are highly heterogeneous in terms of device models and geographical locations. Each sample contains values for different resources, including the battery\_level and battery\_state, at a particular timestamp. We then filter the dataset for high quality traces and then pre-process the data, detailed in Appendix~\ref{app:preproc}.

\subsection{Trace Selection and Re-Sampling} \label{app:preproc}
We observe that the sampling frequencies and sampling periods for users are not consistent given the complicated real-world settings. To utilize the data, we first pre-processed the data. We selected 100 high-quality user traces out of 0.3 million users with the following criteria: 1) The user has a sampling period that is no less than 28 days; 2) The user has an overall sampling frequency no less $\frac{5}{432}$ Hz, which is equivalent to 100 samples one day on average across the whole sampling period; 3) The maximum time gap between two adjacent samples is no larger than 24 hours; 4) The number of time gaps between two adjacent samples that are larger than 6 hours should be no more than 15.  We resample the non-uniform traces using Piecewise Cubic Hermite Interpolating Polynomial (i.e. scipy.interpolate.PchipInterpolator) to a fixed rate of 10min frequency.

After the resampling of "battery\_level", we set the "battery\_state" to reflect whether the battery is charging(1), not discharing(0), or discharging(-1). This depends on the sign of the difference between the current "battery\_level" and the previous "battery\_level" for each pair of consecutive datapoints

\paragraph{Data augmentation for temporal heterogeniety} \label{app:augmentation} In order to simulate client availability across all time zones, we select sub-intervals of 100 traces shifted by 1 hour, 23 times. This results in 2400 clients spread across the planet.

\section{Implementation} \label{app:implementation}

In this section, we discuss the implementation we followed corresponding to each step outlined in Section~\ref{sec:design}.

\paragraph{Scheduling in Android} The first hurdle of running the training process on Android was to ensure that the scheduler does not put the process to sleep once the phone's screen turns off. We solve this issue by acquiring a "WakeLock" \cite{wakelock}, an Android level feature that allows the app unrestricted use of the processor. In order to be able to explore the performance and energy usage of different core combinations, we needed low-level control to limit the scheduling of the training processes to a specfic core or a set of cores and change the number of threads at run-time. This required access to the Linux scheduling API function calls \texttt{sched\_setaffinity} and \texttt{sched\_getaffinity} \cite{linuxaff}. The choice of the deep-learning library implemention thus determines native access to these APIs.

\paragraph{Calculating Energy Cost} The energy is calculated by logging the drop in battery SoC. Instantaneous Power is calculated as Voltage * Current. This can be approximated by averaging the current and voltage over an interval of 1 \% battery level drop. Average Power = $(V_{start} + V_{end}) / 2 * ( battery\_capacity / 100) / \Delta T$, where $V_{start}$ and $V_{end}$ are the battery voltages at the start and end of the interval, and $battery\_capacity$ is the charge capacity of the smartphone's battery in Columbs, and $\Delta T$ being the length of the time interval. The energy can be calculated across every drop in battery level, and thus can be summed up in a piece-wise manner across intervals that overlapped with the benchmark of concern to produce a total energy usage estimate.

\paragraph{Mobile Deep Learning Library} We considered mobile-oriented versions of predominent DL frameworks, like PyTorch and Deeplearning4J(DL4J), since they are already used in client-side executors like KotlinSyft \cite{pysyft}. Although Android's JNI API offers access to system calls, the threading API used by the Android builds of PyTorch and DL4J do not offer a way to change the number of threads at run-time. On the other hand, PyTorch's Linux backend did have a way dynamically control the number of threads when compiled with OpenMP\cite{dagum1998openmp}.

\paragraph{Execution Environment} We utilize a Linux-like environment created by Termux~\cite{termux}, a Linux terminal emulator app running on an Android phone. This lets us access many low-level system calls, including \texttt{sched\_setaffinity} and \texttt{sched\_getaffinity}. We use PyTorch v1.8.1 with OpenMP, compiled on the device for the 64-bit ARM architecture to run the local training. We use the termux-api \cite{termux-api} interface to monotor the system-state, including the battery state and charge level.

\paragraph{Performance and Energy Benchmarking} \label{app:perf-bench} After setting the CPU affinity, training costs associated with a particular deep-learning model is benchmarked by amortizing the battery usage across multiple runs detailed. We then subtract power required to run other processes and components of the phone to arrive at the energy and power usage of the training. In order to minimize the effect of the external environment and other applications on the benchmark, we stop all unneccesary services and background processes to isolate the execution from any interference.

\end{document}